\documentclass[10pt,twocolumn,letterpaper]{article}

\usepackage{cvpr}
\usepackage{times}
\usepackage{epsfig}
\usepackage{graphicx}
\usepackage{amsmath}
\usepackage{amssymb}
\usepackage{amsthm}
\usepackage{dsfont}
\usepackage{algorithm}
\usepackage{algorithmic}
\usepackage{caption}
\usepackage{subcaption}
\usepackage{multirow}

\def\1{\mathds{1}}
\def\D{\Delta}

\def\T{\mathcal{T}}

\def\R{\mathbb{R}}
\def\X{\mathcal{X}}
\def\Y{\mathcal{Y}}

\def\x{\textbf{x}}
\def\wt{\tilde{\textbf{w}}}

\def\y{\textbf{y}}
\def\l{\ell}

\def\ie{i.e\onedot}
\def\eg{e.g\onedot}
\def\cf{cf\onedot}
\def\etc{etc\onedot}
\def\vs{vs\onedot}

\def\train{\texttt{train}}
\def\val{\texttt{val}}
\def\test{\texttt{test}}
\def\mthd{LatEm\xspace}
\def\argmax{\mathop{\rm arg\,max}\limits}%    a math operator.
\newcommand{\myparagraph}[1]{\vspace{5pt}\noindent{\bf #1}}
\graphicspath{{./figures/}}
\newcommand{\cutcaptiondown}{\vspace*{-0.12in}}

\usepackage[pagebackref=true,breaklinks=true,letterpaper=true,colorlinks,allcolors=blue,bookmarks=false]{hyperref}

 \cvprfinalcopy % *** Uncomment this line for the final submission

\def\cvprPaperID{1442} % *** Enter the CVPR Paper ID here

\ifcvprfinal\fi
\begin{document}

%%%%%%%%% TITLE
\title{Latent Embeddings for Zero-shot Classification}

\author{Yongqin Xian$^1$, Zeynep Akata$^1$, Gaurav Sharma$^{1,2,}$\thanks{Currently
with CSE, Indian Institute of Technology Kanpur.  Majority of this work was done at Max Planck Institute for Informatics.}\hspace{1.5mm}, 
Quynh Nguyen$^3$,
Matthias Hein$^3$ and Bernt Schiele$^1$ \vspace{4mm} \\ 
{
\begin{tabular}{cp{0.5cm}cp{0.5cm}c}
$^1$MPI for Informatics & & $^2$IIT Kanpur & & $^3$Saarland University 
\end{tabular}
}
}

\maketitle
\begin{abstract}
We present a novel latent embedding model for learning a compatibility function between image and class embeddings, in the context of zero-shot classification. The proposed method augments the state-of-the-art bilinear compatibility model by incorporating latent variables. Instead of learning a single bilinear map, it learns a collection of maps with the selection, of which map to use, being a latent variable for the current image-class pair. We train the model with a ranking based objective function which penalizes incorrect rankings of the true class for a given image. We empirically demonstrate that our model improves the state-of-the-art for various class embeddings consistently on three challenging publicly available datasets for the zero-shot setting. Moreover, our method leads to visually highly interpretable results with clear clusters of different fine-grained object properties that correspond to different latent variable maps.
\end{abstract}

\section{Introduction}
Zero-shot classification~\cite{HEEY15,LNH13,LEB08,RSS11,YA10} is a challenging problem.  
The task is generally set as follows: training images are provided for certain visual classes and the classifier is expected to predict the presence or absence of novel classes at test time. 
The training and test classes are connected via some auxiliary, non visual source of information \eg attributes. 

\begin{figure}[t]
\begin{center}
\includegraphics[width=\columnwidth, trim=0 20 0 0]{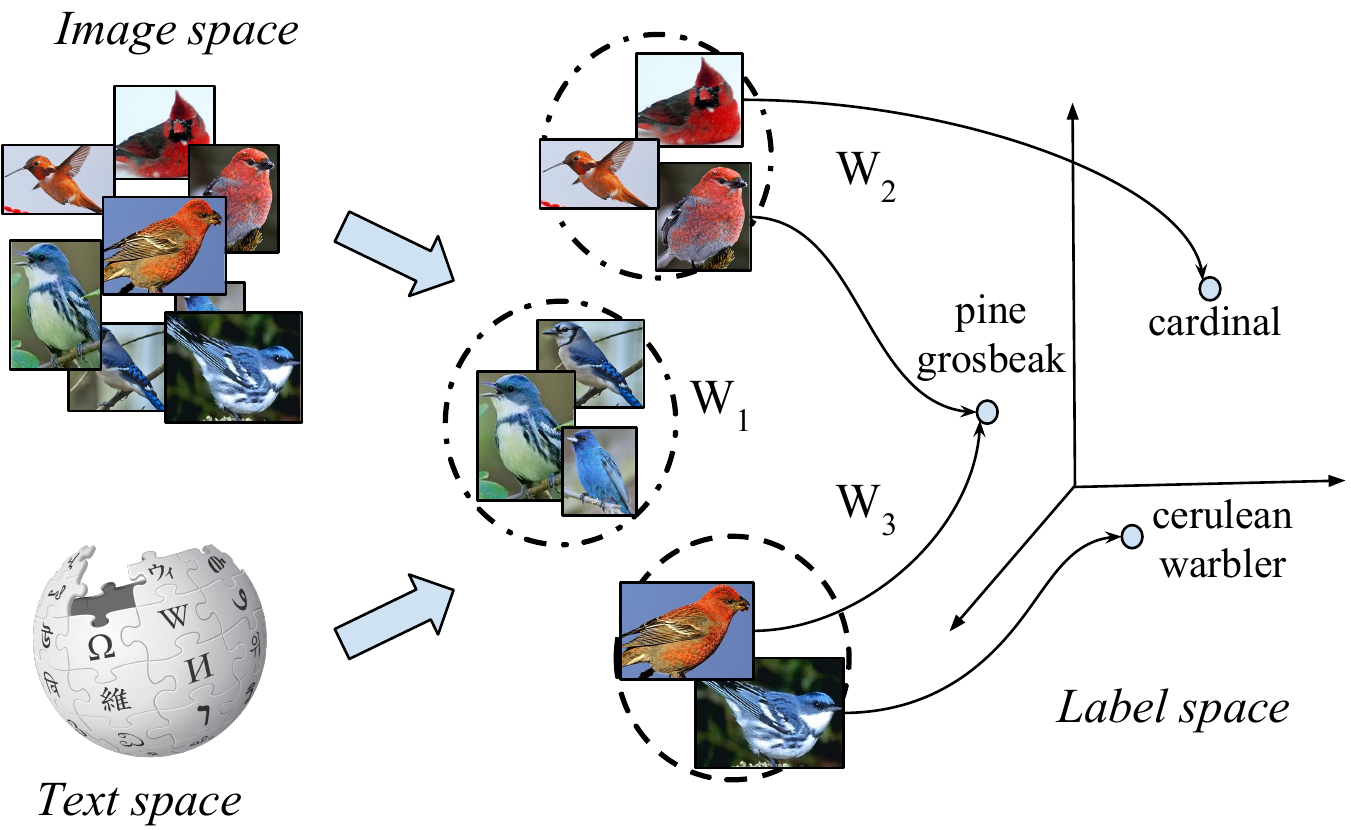}
\end{center}
\caption{\mthd learns multiple $W_i$'s that maximize the compatibility
between the input embedding (image, text space) and the output embedding (label space) of all training examples. 
The different $W_i$'s may capture different visual characteristics of
objects, \ie color, beak shape \etc and allow distribution of the complexity among them, enabling the model to
do better classification.} 
\label{fig:LMM}
\cutcaptiondown
\end{figure}

Combining visual information with attributes~\cite{DPCG12,FEH10,FZ07,KKTH12,LNH13,PG11,PCKF14} has also supported fine grained classification.
In fine grained image collections, images that belong to different classes are visually similar to each other, \eg different bird species.
Image labeling for such collections is a costly process, as it requires either expert opinion or a large number of attributes.
To overcome this limitation, recent works have explored distributed text representations~\cite{MSCCD13,PSM14,WordNet} which are
learned from general (or domain specific) text corpora.

Substantial progress has been made for image classification problem in the zero-shot setting on fine-grained image collections~\cite{ARWLS15}. 
This progress can be attributed to (i) strong deep learning based image features \cite{KSH12, SzegedyCVPR2015}  and (ii) learning a discriminative compatibility function between the structured image and class embeddings~\cite{APHS15,ARWLS15,FCSBM13,romera2015embarrassingly}. 
The focus of this work is on the latter, \ie on improving the compatibility learning framework, in particular via unsupervised auxiliary information.

The main idea of structured embedding frameworks~\cite{APHS15,ARWLS15,FCSBM13,romera2015embarrassingly} is to first represent both the images \emph{and the classes} in some multi-dimensional vector spaces.  
Image embeddings are obtained from state-of-the-art image representations \eg those from convolutional neural networks \cite{KSH12, SzegedyCVPR2015}.  
Class embeddings can either (i) be obtained using manually specified side information \eg
attributes~\cite{LNH13}, or (ii) extracted automatically~\cite{MSCCD13, PSM14} from an unlabeled large text corpora. 
A discriminative bilinear compatibility function is then learned that pulls images from the same class close to each other and pushes images from different classes away from each other. 
Once learned, such a compatibility function can be used to predict the class (more precisely, the embedding) of any given image (embedding).
In particular, this prediction can be done for images from both seen and unseen classes, hence enabling zero-shot classification.

We address the fine-grained zero-shot classification problem 
while being particularly interested in more flexible unsupervised text embeddings. 
The state-of-the-art methods \cite{APHS15,ARWLS15,FCSBM13,romera2015embarrassingly} use a unique, globally linear compatibility function for all types of images.
However, learning a linear compatibility function is not particularly suitable for the challenging fine-grained classification problem.
For fine-grained classification, a model that can automatically group objects with similar properties together
and then learn for each group a separate compatibility model is required.
For instance, two different linear functions that separate blue birds with brown wings and from other blue birds with blue wings 
can be learned separately.
To that end, we propose a novel model for zero-shot setting
which incorporates latent variables to learn a piecewise linear compatibility function between image and class embeddings.
The approach is inspired by many recent advances in visual recognition that
utilize latent variable models, \eg in object detection~\cite{FelzenszwalbPAMI2010,
HussainBMVC2010}, human pose estimation~\cite{YangCVPR2011} and face detection~\cite{Zhu2012face} \etc.

Our contributions are as follows.
(1) We propose a novel method for zero-shot learning. By incorporating latent variables in the compatibility function our method achieves factorization over such (possibly complex combinations of) variations in pose, appearance and other factors. 
Instead of learning a single linear function, we propose to learn a collection of linear models while allowing each image-class pair to choose from them. 
This effectively makes our model non-linear, as in different local regions of the space the decision boundary, while being linear, is different. 
We use an efficient stochastic gradient descent (SGD) based learning method.  
(2) We propose a fast and effective method for model selection, \ie through model pruning.
(3) We evaluate our novel piecewise linear model for zero-shot classification on three challenging datasets. 
We show that incorporating latent variables in the compatibility learning framework consistently improves the state-of-the-art. 

The rest of the paper is structured as follows. In Sec. \ref{sec:bg} we detail the bilinear compatibility learning framework that we base our method on. In Sec. \ref{sec:method} we present our novel Latent Embedding (\mthd) method. In Sec. \ref{sec:experiments} we present our experimental evaluation and in Sec. \ref{sec:conclusions} we conclude.

\section{Related Work}
\label{sec:related}

We are interested in the problem of zero-shot learning where the test classes are disjoint from the training classes~\cite{HEEY15,LNH13,LEB08,RSS11,YA10,ZS16}. 
As visual information from such test classes is not available during training, zero-shot learning requires secondary information sources to make up for the missing visual information. 
While secondary information can come from different sources, usually they are derived from either large and unrestricted, but freely available, text corpora, \eg word2vec~\cite{MSCCD13}, glove~\cite{PSM14}, or structured textual sources \eg wordnet hierarchies~\cite{WordNet}, or costly human annotations \eg manually specified attributes~\cite{DPCG12,FEH10,FZ07,KKTH12,LNH13,PG11,PCKF14}. 
Attributes, such as `furry', `has four legs' \etc for animals, capture several characteristics of objects (visual classes) that help associate some and differentiate others. 
They are typically collected through costly human annotation~\cite{DPCG12, KKTH12, PG11} and have shown promising results~\cite{APHS15,CGG13,DRS11,LNH13,LKS11,SKBB12,SFD11,YJKLGL11} in various computer vision problems.

The image classification problem, with a secondary stream of information, could be either solved by
solving related sub-problems, \eg attribute prediction~\cite{LNH13,RSS11,RSS10}, or by a direct
approach, \eg compatibility learning between embeddings~\cite{APHS15,FCSBM13,WBU11}. One such
factorization could be by building intermediate attribute classifiers and then making a class
prediction using a probabilistic weight of each attribute for each sample \cite{LNH13}. However,
these methods, based on attribute classifiers, have been shown to be suboptimal~\cite{APHS15}. This is due to their reliance on binary mappings (by
thresholding attribute scores) between attributes and images which causes loss in information.
On the other hand, solving the problem directly, by learning a direct mapping between images and
their classes (represented as numerical vectors) has been shown to be better suited. 
Such label
embedding methods~\cite{APHS15,ARWLS15,FCSBM13,Hastie:Tibshirani:Friedman:2008,NMBSSFCD13,palatucci2009zero,romera2015embarrassingly,socher2013zero} aim to
find a mapping between two embedding spaces, one each for the two streams of information \eg visual
and textual. Among these methods, CCA~\cite{Hastie:Tibshirani:Friedman:2008} maximizes the
correlation between these two embedding spaces, ~\cite{palatucci2009zero} learns a linear compatibility between an fMRI-based image space and the semantic space, ~\cite{socher2013zero} learns a deep non-linear mapping between images and tags, ConSe~\cite{NMBSSFCD13} uses the probabilities of a softmax-output layer to weight the vectors of all the classes, SJE~\cite{ARWLS15} and ALE~\cite{APHS15} learn a bilinear compatibility function using a multiclass~\cite{CS02} and a weighted
approximate ranking loss~\cite{Jo06} respectively. DeViSE~\cite{FCSBM13} does the same, however,
with an efficient ranking formulation. Most recently, \cite{romera2015embarrassingly} proposes to learn this mapping by optimizing a simple objective function which has closed form solution. 

We build our work on multimodal embedding methods. 
However, instead of learning a linear compatibility function, we propose a nonlinear compatibility framework 
that learns a collection of such linear models making the overall function piecewise
linear.

\section{Background: Bilinear Joint Embeddings}
\label{sec:bg}

In this section, we describe the bilinear joint embedding framework~\cite{WBU11,APHS15,ARWLS15}, on which we build our Latent Embedding Model (\mthd)
(Sec.~\ref{sec:method}).

We work in a supervised setting where we are given an annotated training set $\T = \{(\x,\y)| \x \in \X \subset \R^{d_x}, \y \in \Y \subset \R^{d_y}\}$
where $\x$ is the image embedding defined in an image feature space $\X$, \eg CNN
features~\cite{KSH12}, and $\y$ is the class embedding defined in a label space $\Y$ that models the
conceptual relationships between classes, \eg attributes~\cite{FEHF09,LNH13}.
The goal is to learn a function $f:\X\rightarrow \Y$ to predict the correct class for the query images. 
In previous work \cite{WBU11,APHS15,ARWLS15}, this is done via
learning a function $F:\X \times \Y \rightarrow \R$ 
that measures the compatibility between a given input embedding ($\x\in\X$) and an output embedding ($\y\in\Y$). 
The prediction function then chooses the class with the maximum compatibility, \ie
\begin{align} 
f(\x) = \arg\max_{\y \in \Y} F(\x,\y).
\end{align}
In general, the class embeddings reflect the common and distinguishing properties of different classes using
side-information that is extracted independently of images. 
Using these embeddings, the compatibility can be computed even with those unknown classes which have no corresponding images 
in the training set.
Therefore, this framework can be applied to zero-shot learning \cite{APHS15,ARWLS15,palatucci2009zero,romera2015embarrassingly,socher2013zero}. 
In previous work, the compatibility function takes a simple form,
\begin{align}
\label{eqnLinComp}
F(\x,\y) = \x^\top W \y
\end{align}
with the matrix $W \in \R^{d_x \times d_y}$ being the parameter to be learnt from training data. 
Due to the bilinearity of $F$ in $\x$ and $\y$, previous work \cite{APHS15,ARWLS15,WBU11} refer to this model as a bilinear model,
however one can also view it as a linear one since $F$ is linear in the parameter $W.$
In the following, these two terminologies will be used interchangeably depending on the context.

\section{Latent Embeddings Model (\mthd)}
\label{sec:method}

In general, the linearity of the compatibility function (Eq.~\eqref{eqnLinComp}) is a limitation
as the problem of image classification is usually a complex nonlinear decision problem.  A very successful
extension of linear decision functions to nonlinear ones, has been through the use of
piecewise linear decision functions. This idea has been applied successfully to various computer
vision tasks \eg mixture of templates \cite{HussainBMVC2010} and deformable parts-based model
\cite{FelzenszwalbPAMI2010} for object detection, mixture of parts for pose estimation \cite{YangCVPR2011} and face
detection \cite{Zhu2012face}. The main idea in most of such models, along with modeling parts, is
that of incorporating latent variables, thus making the decision function piecewise linear, \eg the
different templates in the mixture of templates \cite{HussainBMVC2010} and the different
`components' in the deformable parts model \cite{FelzenszwalbPAMI2010}. The model then becomes a
collection of linear models and the test images pick one from these linear models, with the
selection being latent. Intuitively, this factorizes the decision function into components which focus on
distinctive `clusters' in the data \eg one component may focus on the profile view while another on
the frontal view of the object.

\myparagraph{Objective.}
We propose to construct a nonlinear, albeit piecewise linear, compatibility function.  Parallel to
the latent SVM formulation, we propose a non-linear compatibility function as follows,
\begin{align}
\label{eqnLatComp}
F(\x,\y) = \max_{1\leq i \leq K} \wt_i^\top (\x \otimes \y),
\end{align}
where $i=1,\ldots,K$, with $K\geq2$, indexes over the latent choices and $\wt_i \in \R^{d_xd_y}$ are
the parameters of the individual linear components of the model. This can be rewritten as a
mixture of bilinear compatibility functions from Eq.~\eqref{eqnLinComp} as
\begin{align}\label{eqnLatCompBL}
F(\x,\y) = \max_{1\leq i \leq K} \x^\top W_i \y .
\end{align}
Our main goal is to learn a set of compatibility spaces that minimizes the following empirical risk,
\begin{equation}
\frac{1}{N} \sum_{n=1}^{|\T|} L(\x_n, \y_n)  ,
\label{eqn:wsabie}
\end{equation}
where $L: \X\times\Y\to \R$ is the loss function defined for a particular example $(\x_n,\y_n)$ as
\begin{equation}\label{eqnRankLoss}
    L(\x_n,\y_n) = \sum_{\y \in \Y} \max\{0,  \D(\y_n,\y) + F(\x_n,\y) -F(\x_n,\y_n)   \} 
\end{equation} 
where $\D(\y,\y_n)=1$ if $\y\neq\y_n$ and $0$ otherwise.
This ranking-based loss function has been previously used in \cite{FCSBM13, WBU11} 
such that the model is trained to produce a higher compatibility between the image embedding and the class embedding of the
correct label than between the image embedding and class embedding of other labels.

\begin{algorithm}[t]
  \caption{SGD optimization for LatEm}
   $\T = \{(\x,\y)| \x \in \R^{d_x}, \y \in \R^{d_y}\}$
  \begin{algorithmic}[1]
    \FORALL {$t=1$ to $T$} 
	\FORALL {$n=1$ to $|\T|$} 
      		\STATE  Draw $(\x_n$,$\y_n) \in \T$ and $\y \in \Y \setminus \{\y_n\}$ 
      			\IF {$F(\x_n,\y) + 1 > F(\x_n,\y_n)$} 
				\STATE $i^* \leftarrow \argmax\limits_{1\leq k\leq K} \x_n^\top W_k \y$
				\STATE $j^* \leftarrow \argmax\limits_{1\leq k\leq K} \x_n^\top W_k \y_n $  
				\IF {$ i^* = j^*$} 
				\STATE $W_{i^*}^{t+1} \leftarrow W_{i^*}^{t} - \eta_t \x_n (\y - \y_n)^\top$
				\ENDIF
				\IF {$ i^* \neq j^*$} 
          			\STATE $W_{i^*}^{t+1} \leftarrow W_{i^*}^{t} - \eta_t \x_n \y^\top$ 
          			\STATE $W_{j^*}^{t+1} \leftarrow W_{j^*}^{t} + \eta_t \x_n \y_n^\top$ 
          			\ENDIF
        		\ENDIF
      \ENDFOR
    \ENDFOR
  \end{algorithmic}
  \label{alg:sgd}
\end{algorithm} 

\myparagraph{Optimization.}
To minimize the empirical risk in Eq.~\eqref{eqn:wsabie}, 
one first observes that the ranking loss function $L$ from Eq.~\eqref{eqnRankLoss} is not jointly convex in all the $W_i$'s 
even though $F$ is convex.
Thus, finding a globally optimal solution as in the previous linear models \cite{APHS15,ARWLS15} is out of reach.
To solve this problem, we propose a simple SGD-based method that works in the same fashion as in the convex setting.
It turns out that our algorithm works well in practice and achieves state-of-the-art results 
as we empirically show in Sec. \ref{sec:experiments}.

We explain the details of our Algorithm \ref{alg:sgd} as follows.
We loop through all our samples for a certain number of epochs $T$.
For each sample $(\x_n,\y_n)$ in the training set, we randomly select a $\y$ that is different from $\y_n$ (step $3$ of Algorithm \ref{alg:sgd}).
If the randomly selected $\y$ violates the margin (step $4$ in Algorithm \ref{alg:sgd}), 
then we update the $W_i$ matrices (steps $5-13$ in Algorithm \ref{alg:sgd}). 
In particular, we find the $W_i$ that leads to the maximum score for $\y$ and the $W_j$ that gives the maximum score for $\y$.
If the same matrix gives the maximum score (step $7$ in Algorithm \ref{alg:sgd}), we update that matrix. 
If two different matrices lead to the maximum score (step $9$ in Algorithm \ref{alg:sgd}), we update them both using SGD.

\myparagraph{Model selection.}
The number of matrices $K$ in the model is a free parameter. We use two strategies to select the
number of matrices. As the first method, we use a standard cross-validation strategy -- we split the
dataset randomly into disjoint parts (in a zero-shot setup) and choose the $K$ with the best
cross-validation performance. While this is a well established strategy which we find to work well
experimentally, we also propose a pruning based strategy which is competitive while being faster to
train. As the second method, we start with a large number of matrices and prune them as
follows. As the training proceeds, each sampled training examples chooses one of the matrices for
scoring -- we keep track of this information and build a histogram over the number of matrices
counting how many times each matrix was chosen by any training example. 
In particular, this is done by increasing the counter for $W_{j^*}$ by $1$ after step $6$ of Algorithm \ref{alg:sgd}.
With this information,
after five passes over the training data, we prune out the matrices which were chosen
by less than $5\%$ of the training examples, so far. This is based on the
intuition that if a matrix is being chosen only by a very small number of examples, it is probably
not critical for performance. With this approach we have to train only one model which adapts
itself, instead of training multiple models for cross-validating $K$ and then training a final model
with the chosen $K$.

\myparagraph{Discussion.} \mthd builds on the idea of Structured Joint Embeddings (SJE)~\cite{ARWLS15}. 
We discuss below the differences between \mthd and SJE and emphasize our technical contributions. 

\mthd learns a piecewise linear compatibility function through multiple $W_i$ matrices whereas SJE~\cite{ARWLS15} is linear.
With multiple $W_i$'s the compatibility function has the freedom to treat different types of images differently. 
Let us consider a fixed class $\hat{\y}$ and two substantially visually different types of images $\x_1, \x_2$, \eg the same bird flying and swimming. 
In SJE~\cite{ARWLS15} these images will be mapped to the class embedding space with a single mapping $W^\top \x_1, W^\top \x_2$.
On the other hand, \mthd will have learned two different matrices for the mapping \ie $W_1^\top \x_1, W_2^\top \x_2$.  
While in the former case a single $W$ has to map two visually, and hence numerically, very different vectors (close) to the
same point, in the latent case such two different mappings are factorized separately and hence are
arguably easier to perform. 
Such factorization is also expected to be advantageous when two classes sharing partial visual similarity are to be discriminated \eg while blue birds could be relative easily distinguished from red birds, to do so for different types of blue birds is harder. 
In such cases, one of the $W_i$'s could focus on color while another one could focus on the beak shape (in Sec. \ref{subsec:qualitative} we show that this effect is visible).
The
task of discrimination against different bird species would then be handled only by the second one, which would also
arguably be easier.

\mthd uses the ranking based loss~\cite{WBU11} in Eq.~\eqref{eqnRankLoss} whereas SJE~\cite{ARWLS15} uses the multiclass loss of Crammer and Singer~\cite{CS02} which replaces the $\sum$ in Eq.~\eqref{eqnRankLoss} with $\max$. The SGD algorithm for multiclass loss of Crammer and Singer~\cite{CS02} requires at each iteration a full pass over all the classes to search for the maximum violating class. Therefore it can happen that some matrices will not be updated frequently. On the other hand, the ranking based loss in Eq.~\eqref{eqnRankLoss} used by our LatEm model ensures that different latent matrices are updated frequently. Thus, the ranking based loss in Eq.~\eqref{eqnRankLoss} is better suited for our piecewise linear model.

\section{Experiments}
\label{sec:experiments}
 
We evaluate the proposed model on three challenging publicly available datasets of Birds, Dogs and
Animals. First, we describe the
datasets, then give the implementation details and finally report the experimental results.

 \begin{table}[t]
 	\begin{center}
 		\begin{tabular}{|c ||c |c ||c |c || c | c |}
 			\hline
 			& \multicolumn{2}{c||}{Total} & \multicolumn{2}{|c||}{\train+\val} & \multicolumn{2}{|c|}{\test} \\
 			    & imgs  & cls & imgs  & cls & imgs  & cls \\ \hline \hline
 			{CUB} & $11786$  & $200$   & $8855$   & $150$   & $2931$   & $50$      \\ \hline
	 		{AWA} & $30473$  & $50 $   & $24293$  & $40$    & $6180$   & $10$ \\ \hline
 			{Dogs} & $19499$  & $113$   & $14681$  & $85$    & $4818$   & $28$ \\ \hline
 		\end{tabular}
 		
 		\caption{The statistics of the three datasets used. CUB and Dog are fine-grained datasets whereas AWA is a more general concept dataset.}
        \label{tabDBStats}
 	\end{center}
 \end{table}

\myparagraph{Datasets.} Caltech-UCSD Birds (CUB), Stanford Dogs (Dogs) are standard benchmarks of fine-grained recognition~\cite{DPCG12,DKF13,CaltechUCSDBirdsDataset,StanfordDogsDataset} and  Animals With Attributes (AWA) is another popular and
challenging benchmark dataset~\cite{LNH13}. All these three datasets have been used for zero-shot learning~\cite{ARWLS15,RSS11,KKTH12,YA10}. 
Tab.~\ref{tabDBStats} gives the statistics for them. 

In zero-shot setting, the dataset is divided into three disjoint sets of \train, \val\ and \test.
For comparing with previous works, we follow the same \train/\val/\test\ set split used
by~\cite{ARWLS15}. In zero-shot learning, where training and test classes are disjoint sets, 
to get a more stable estimate in our own results, we make four more splits by random
sampling, while keeping the number of classes the same as before.
We average results over the total of five splits. The average performance over the five splits is
the default setting reported in all experiments, except where mentioned otherwise, \eg in comparison
with previous methods.

\myparagraph{Image and class embeddings.} In our latent embedding (\mthd) model, the image
embeddings (image features) and class embeddings (side information) are two essential
components. To facilitate direct comparison with the state-of-the-art, we use the embeddings
provided by~\cite{ARWLS15}. Briefly, as image embeddings we use the $1,024$ dimensional
outputs of the top-layer pooling units of the pre-trained GoogleNet~\cite{SzegedyCVPR2015} extracted
from the whole image. We do not do any task specific pre-processing on images such as cropping foreground objects.

As class embeddings we evaluate four different alternatives, \ie attributes (\texttt{att}), word2vec (\texttt{w2v}),
glove (\texttt{glo}) and hierarchies (\texttt{hie}). Attributes~\cite{LNH13,FEHF09} are distinguishing properties of
objects that are obtained through human annotation. For fine-grained datasets such as CUB and Dogs,
as objects are visually very similar to each other, a large number of attributes are needed. Among
the three datasets used, CUB contains $312$ attributes, AWA contains $85$ attributes while Dogs does
not contain annotations for attributes. Our attribute class embedding is a vector per-class measuring
the strength of each attribute based on human judgment.

In addition to human annotation the class embeddings can be constructed automatically from either
a large unlabeled text corpora or through hierarchical relationship between classes. This has certain 
advantages such as we do not need any costly human annotation, however as a drawback, they tend
not to perform as well as supervised attributes. One of our motivations for this work is that
the class embeddings captured from a large text corpora contains latent relationships between 
classes and we would like to automatically learn these. Therefore, 
we evaluate three common methods for building unsupervised text embeddings. Word2Vec~\cite{MSCCD13}
is a two-layer neural network which predicts words given the context within a skip window slided
through a text document. It builds a vector for each word in a learned vocabulary.
Glove~\cite{PSM14} is another distributed text representation method which uses co-occurrence
statistics of words within a document. We use the pre-extracted word2vec and glove vectors from
\texttt{wikipedia} provided by~\cite{ARWLS15}. Finally, another way of building a vectorial
structure for our classes is to use a hierarchy such as WordNet~\cite{WordNet}. Our hierarchy
vectors are based on the hierarchical distance between child and ancestor nodes, in WordNet,
corresponding to our class names. For a direct comparison, we again use the hierarchy vectors
provided by~\cite{ARWLS15}. In terms of size, \texttt{w2v} and \texttt{glo} are $400$ dimensional whereas \texttt{hie} is 
$\approx 200$ dimensional.

\myparagraph{Implementation details.} Our image features are z-score normalized such that each
dimension has zero mean and unit variance. All the class embeddings are $\l_2$ normalized.  The
matrices $W_i$ are initialized at random with zero mean and standard deviation
$\frac{1}{\sqrt{d_x}}$~\cite{APHS15}. The  number of epochs is fixed to be $150$. The learning
rates for the CUB, AWA and Dog datasets are chosen as $\eta_t=0.1, 0.001, 0.01$, respectively, and kept
constant over iterations.  For each dataset, these parameters are tuned on the validation set of the 
default dataset split and kept constant for all other dataset folds and for all class embeddings.
As discussed in Sec \ref{sec:method}, we perform two strategies for selecting the number of latent matrices $K$: 
cross-validation and pruning.
When using cross-validation, $K$ is varied in $\{2, 4, 6, 8, 10\}$ and 
the optimal $K$ is chosen based the accuracy on a validation set.
For pruning, $K$ is initially set to be $16,$ and then at every fifth epoch during training, 
we prune all those matrices that support less than $5\%$ of the data points.

    \begin{table}[t]
    	\centering
        \resizebox{\columnwidth}{!} {
        \begin{tabular}{|c||c|c||c|c||c|c|}
            \hline
            & \multicolumn{2}{c||}{CUB} & \multicolumn{2}{c||}{AWA} & \multicolumn{2}{c|}{Dogs} \\ 
            & SJE  & \mthd & SJE  & \mthd & SJE  & \mthd \\ \hline \hline
            \texttt{att} &$\mathbf{50.1}$ & $45.5$ & $66.7$ & $\mathbf{71.9}$ &  N/A & N/A    \\ \hline
            \texttt{w2v} &$28.4$ & $\mathbf{31.8}$ &  $51.2$ & $\mathbf{61.1}$  & $19.6$ & $\mathbf{22.6}$ \\ \hline
            \texttt{glo} &$24.2$ & $\textbf{32.5}$  & $58.8$ & $\mathbf{62.9}$  & $17.8$ & $\textbf{20.9}$  \\ \hline
            \texttt{hie} &$20.6$ & $\textbf{24.2}$ & $51.2$ & $\mathbf{57.5}$ & $24.3$ & $\mathbf{25.2}$  \\ \hline
        \end{tabular}
        }
       \caption{Comparison of Latent Embeddings (\mthd) method with the state-of-the-art SJE~\cite{ARWLS15} method. 
       We report average per-class Top-1 accuracy on unseen classes.
       We use the same data partitioning, same image features and same class embeddings as SJE~\cite{ARWLS15}.
       We cross-validate the $K$ for \mthd.}
       \label{tab:soa}
    \end{table}

\begin{table}[t]
  \centering 
  \resizebox{\columnwidth}{!} {
    \begin{tabular}{|c||c|c||c|c||c|c|}
    \hline
    & \multicolumn{2}{c||}{CUB} & \multicolumn{2}{c||}{AWA} & \multicolumn{2}{c|}{Dogs} \\ 
     & SJE  & \mthd & SJE  & \mthd & SJE  & \mthd \\ \hline \hline
    w  & $\mathbf{51.7}$ & $47.4$ & $73.9$ & $\mathbf{76.1}$ & N/A & N/A \\
    \hline
    w/o & $29.9$ & $\mathbf{34.9}$ & $60.1$ & $\mathbf{66.2}$ & $35.1$ & $\mathbf{36.3}$ \\       
    \hline           
    \end{tabular}
    }
  \caption{Combining embeddings either including or not including supervision in the combination. w: the combination includes attributes, w/o: the combination does not include attributes.}
  \label{tab:cmb}
\end{table}

\subsection{Comparison with State-of-the-Art}
\label{subsec:soa}
We now provide a direct comparison between our \mthd and the state-of-the-art SJE~\cite{ARWLS15} method.
SJE (Sec.~\ref{sec:bg}) learns a bilinear function that maximizes the  compatibility between image and class embeddings.
Our \mthd on the other hand learns a nonlinear, \ie piece-wise linear function, through multiple compatibility functions defined between image and class embeddings.

The results are presented in Tab.~\ref{tab:soa}. Using the text embeddings obtained through human annotation, \ie attributes (\texttt{att}), \mthd improves over SJE on AWA ($71.9\%$ \vs $66.7\%$) significantly.
However, as our aim is to reduce the accuracy gap between supervised and unsupervised class
embeddings, we focus on unsupervised embeddings, \ie \texttt{w2v}, \texttt{glo} and \texttt{hie}.  
On all datasets, \mthd with \texttt{w2v}, \texttt{glo} and \texttt{hie} improves the state-of-the-art SJE~\cite{ARWLS15} significantly. 
With \texttt{w2v}, \mthd achieves $31.8\%$ accuracy (vs $28.4\%$) on CUB, $61.1\%$ accuracy (vs $51.2\%$) on AWA and finally $22.6\%$ (vs $19.6\%$) on Dogs. 
Similarly, using \texttt{glo}, \mthd achieves $32.5\%$ accuracy (vs $24.2\%$) on CUB, $62.9\%$ accuracy (\vs $58.8\%$) on AWA and $20.9\%$ accuracy (\vs $17.8\%$) on Dogs. 
Finally, while \mthd with \texttt{hie} on Dogs improves the result to $25.2\%$ from $24.3\%$, the improvement is more significant on CUB ($24.2\%$ from $20.6\%$) and on AWA ($57.5\%$ from $51.2\%$). 
These results establish our novel Latent Embeddings (\mthd) as the new state-of-the-art method for zero-shot learning on three datasets in ten out of eleven test settings. 
They are encouraging, as they quantitatively show that learning piecewise linear latent embeddings indeed capture latent semantics on the class embedding space. 

Following~\cite{ARWLS15} we also include a comparison when combining supervised and unsupervised embeddings.
The results are given in Tab~\ref{tab:cmb}.
First, we combine all the embeddings, \ie \texttt{att,w2v,glo,hie} for AWA and CUB. 
\mthd improves the results over SJE significantly on AWA ($76.1\%$ vs $73.9\%$).
Second, we combine the unsupervised class embeddings, \ie \texttt{w2v,glo,hie}, for all datasets.
\mthd consistently improves over the combined embeddings obtained with SJE in this setting.
On CUB combining \texttt{w2v,glo,hie} achieves $34.9\%$ (vs $29.9\%$), on AWA, it achieves $66.2\%$ (vs $60.1\%$) and on Dogs, it obtains $36.3\%$ (vs $35.1\%$).
These experiments show that the embeddings contain non-redundant information, therefore the results tend to improve by combining them.

\begin{table}[t]
    \begin{center}
        \resizebox{\columnwidth}{!} {
        \begin{tabular}{|c ||c |c ||c |c || c | c |}
            \hline
            & \multicolumn{2}{c||}{CUB} & \multicolumn{2}{c||}{AWA} & \multicolumn{2}{c|}{Dogs} \\ 
            & SJE  & \mthd & SJE  & \mthd & SJE  & \mthd  \\ \hline \hline
        \texttt{att} & $\mathbf{49.5}$ & $45.6 $ & $ 70.7 $ & $ \mathbf{72.5} $ &  N/A  &  N/A    \\ \hline
        \texttt{w2v} & $27.7 $ & $ \mathbf{33.1} $ & $ 49.3 $ & $ \mathbf{52.3} $ & $ 23.0 $ & $ \mathbf{24.5}$ \\ \hline
        \texttt{glo}	& $24.8 $ & $ \mathbf{30.7} $ & $ 50.1 $ & $ \mathbf{50.7} $ & $ 14.8 $ & $ \mathbf{20.2}$ \\ \hline
        \texttt{hie} & $21.4 $ & $ \mathbf{23.7} $ & $ 43.4 $ & $ \mathbf{46.2} $ & $ 24.6 $ & $ \mathbf{25.6}$ \\ \hline
        \end{tabular}
        }
       \caption{Average per-class top-1 accuracy on unseen classes (the results are averaged on
       five folds). SJE:~\cite{ARWLS15}, \mthd: Latent embedding model ($K$ is cross-validated). }
       \label{tab:5fold}
       \end{center}       
\end{table}

\myparagraph{Stability evaluation of zero-shot learning.}
Zero-shot learning is a challenging problem due to the lack of labeled training data. 
In other words, during training time, neither images nor class relationships of 
test classes are seen. As a consequence, zero-shot learning suffers from the difficulty in 
parameter selection on a zero-shot set-up, \ie \train, \val\ and \test\ classes
belong to disjoint sets. In order to get stable estimates of our predictions, we experimented on
additional (in our case four) independently and randomly chosen data splits in addition to the standard one. Both with
our \mthd and the publicly available implementation of SJE~\cite{ARWLS15} we repeated the
experiments five times.

The results are presented in Tab~\ref{tab:5fold}.
For all datasets, all the result comparisons between SJE and \mthd hold and therefore the conclusions are the same. 
Although the SJE outperforms \mthd with supervised attributes on CUB, \mthd outperforms the SJE results with supervised attributes on AWA and consistently outperforms all the SJE results obtained with unsupervised class embeddings.
The details of our results are as follows.
Using supervised class embeddings, \ie attributes, on AWA, \mthd obtains an impressive $72.5\%$ (vs $70.5\%$) and using unsupervised embeddings the highest accuracy is observed with \texttt{w2v} with $52.3\%$ (vs $49.3\%$).
On CUB, \mthd with \texttt{w2v} obtains the highest accuracy among the unsupervised class embeddings with $33.1\%$ (vs $27.7\%$)
On Dogs, \mthd with \texttt{hie} obtains the highest accuracy among all the class embeddings, \ie $25.6\%$ (vs $24.6\%$).
These results insure that our accuracy improvements reported in Tab~\ref{tab:soa} were not due to a dataset bias. 
By augmenting the datasets with four more splits, our \mthd obtains a consistent improvement on all the class embeddings on all datasets over the state-of-the-art.

\begin{figure*}[t]
	\centering
	\includegraphics[width=\linewidth]{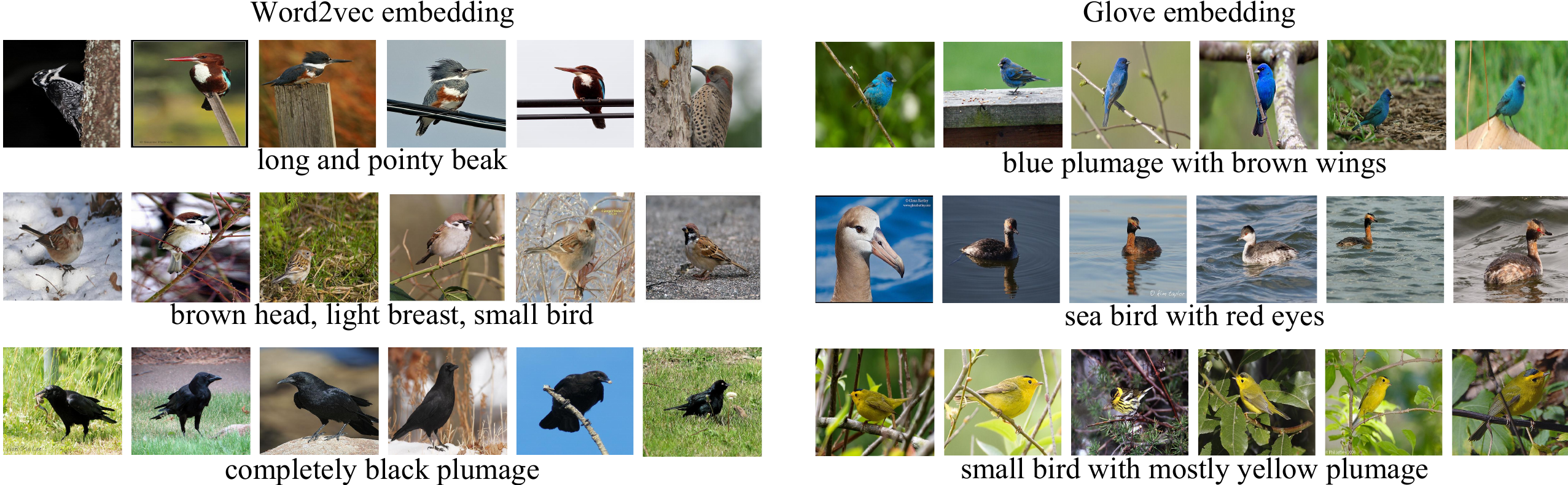}
    \caption{Top images ranked by the matrices using word2vec and glove on CUB dataset, each row
    corresponds to different matrix in the model. Qualitative examples support our intuition -- each
    latent variable captures certain visual aspects of the bird. Note that, while the images may not belong
    to the same fine-grained class, they share common visual properties.} 
	\label{fig:qualitative}
	\cutcaptiondown
\end{figure*}

Note that, for completion, in this
section we provided a full comparison with the state-of-the-art on all class embeddings, including
supervised attributes. However, there are two disadvantages of using attributes.  First, since fine-grained 
object classes share many common properties, we need a large number of attributes
which is costly to obtain.  Second, attribute annotations need to be done on a dataset basis, \ie
the attributes collected for birds do not work with dogs.  Consequently, attribute based methods are
not generalizable across datasets.  Therefore, we are interested in the unsupervised text embeddings
settings, \ie \texttt{w2v}, \texttt{glo}, \texttt{hie}.  
Moreover, with these unsupervised embeddings, our \mthd outperforms the SJE on nine out of nine cases in all our datasets.
For the following sections, we will present results only with \texttt{w2v}, \texttt{glo} and \texttt{hie}.

\subsection{Interpretability of latent embeddings}
\label{subsec:qualitative}

In Sec. \ref{subsec:soa}, we have demonstrated that our novel latent embedding method improves the
state-of-the-art of zero-shot classification
on two fine-grained datasets of birds and dogs, \ie CUB and Dogs, and one dataset of Animals, \ie AWA.
In this section, we zoom into the challenging CUB dataset and aim to investigate if individual $W_i$'s
learn visually consistent and interpretable latent relationships between images and classes.  We
use word2vec and glove as text embeddings. Fig~\ref{fig:qualitative} shows the top scoring images retrieved by three different $W_i$ for the two embeddings \ie \texttt{w2v} and \texttt{glo}.  

For \texttt{w2v}, we observe that the images scored highly by the same $W_i$ (each row)
share some visual aspect. The images in the first row are consistently of birds which have long and
pointy beaks. Note that they belong to different classes; having a long and pointy beak is one of the
shared aspect of birds of these classes. Similarly, for the second row, the retrieved images are of
small birds with brown heads and light colored breasts and the last row contains large birds with 
completely black plumage. These results are interesting because although \texttt{w2v} is trained
on wikipedia in an unsupervised manner with no notion of attributes, our \mthd is able to 
(1) infer hidden common properties of classes and (2) support them with visual evidence, leading to a 
data clustering which is optimized for classification, however also performs well in retrieval.

For \texttt{glo}, similar to the results with \texttt{w2v}, the top-scoring images using the same
$W_i$ consistently show distinguishing visual properties of classes. 
The first row shows blue birds although belonging to different species, are clustered together which indicates
that this matrix captures the ``blue"ness of the objects.  The second row
has exclusively aquatic birds, surrounded by water.  Finally, the third row has yellow
birds only.
Similar to \texttt{w2v}, although \texttt{glo} is trained in an unsupervised manner, our \mthd is able to bring out the 
latent information that reflect object attributes and support this with its visual counterpart.

These results clearly demonstrate that our model factorizes the information with visually
interpretable relations between classes. 

\begin{table*}
\centering
\begin{minipage}{0.47\textwidth}
	\begin{center}
        {
		\begin{tabular}{|c ||c |c ||c |c || c | c |}
			\hline
			& \multicolumn{2}{c||}{CUB} & \multicolumn{2}{c||}{AWA} & \multicolumn{2}{c|}{Dogs} \\ 
			& PR  & CV & PR  & CV & PR  & CV  \\ \hline \hline
			\texttt{att} & $3 $ & $ 4 $ & $ 7 $ & $ 2 $ &   N/A    &  N/A     \\ \hline
			\texttt{w2v} & $8 $ & $ 10 $ & $ 8 $ & $ 4 $ & $ 6  $ & $ 8$ \\ \hline
			\texttt{glo}	& $6 $ & $ 10 $ & $ 7 $ & $ 6 $ & $ 9  $ & $ 4$ \\ \hline
			\texttt{hie} & $8 $ & $ 2  $ & $ 7 $ & $ 2 $ & $ 11 $ & $ 10$ \\ \hline
		\end{tabular}
		}
        \end{center}
\end{minipage}
\begin{minipage}{0.47\textwidth}
	\begin{center}
        {
		\begin{tabular}{|c ||c |c ||c |c || c | c |}
			\hline
			& \multicolumn{2}{c||}{CUB} & \multicolumn{2}{c||}{AWA} & \multicolumn{2}{c|}{Dogs} \\ 
			& PR  & CV & PR  & CV & PR  & CV  \\ \hline \hline
			\texttt{att} & $43.8$ & $\mathbf{45.6} $ & $ 63.2 $ & $ \mathbf{72.5} $ &   N/A    &   N/A   \\ \hline
			\texttt{w2v} & $\mathbf{33.9}$ & $ 33.1 $ & $ 48.9 $ & $ \mathbf{52.3} $ & $ \mathbf{25.0} $ & $ 24.5$ \\ \hline
			\texttt{glo}	& $\mathbf{31.5} $ & $ 30.7 $ & $ \mathbf{51.6} $ & $ 50.7 $ & $ 18.8 $ & $ \mathbf{20.2}$ \\ \hline
			\texttt{hie} & $\mathbf{23.8} $ & $ 23.7 $ & $ 45.5 $ & $ \mathbf{46.2} $ & $ 25.2 $ & $ \mathbf{25.6}$ \\ \hline
		\end{tabular}
		}
        \end{center}
\end{minipage}
\caption{(Left) Number of matrices selected (on the original split) and (right) average per-class
top-1 accuracy on unseen classes (averaged over five splits).   PR: proposed
model learnt with pruning, CV: with cross validation. }
\label{tabPruCV}
\end{table*}

\subsection{Pruning \vs cross-validation for model selection}
\label{subsec:prune}
In this section we evaluate the performances obtained with the number of matrices in the model is
fixed with pruning \vs cross-validation. 

Tab.~\ref{tabPruCV} presents the number of matrices
selected by two methods along with their performances on three datasets. In terms of
performance, both methods are competitive. Pruning outperforms cross validation on five cases and is
outperformed on the remaining six cases. The performance gaps are usually within 1-2\% absolute, with the
exception of AWA dataset with \texttt{att} and \texttt{w2v} with $72.5\%$ \vs $70.7\%$ and $52.3\%$ \vs $49.3\%$,
respectively for cross validation and pruning. Hence neither of the methods has a clear advantage
in terms of performance, however cross validation is slightly better.

In terms of the model size, cross validation seems to have a slight advantage. It selects a smaller model,
hence more space and time efficient one, seven cases out of eleven. The trend is consistent for all class
embeddings for the AwA dataset but is mixed for CUB and Dogs. The advantage of pruning over cross-validation
 is that it is much faster to
train -- while cross validation requires training and testing with multiple models (once each per
every possible choice of $K$), pruning just requires training once. There is however another free
parameter in pruning \ie choice of the amount of training data supporting a matrix for it
to survive pruning. Arguably, it is more intuitive than setting directly the number of matrices to
use instead of cross validating.

\begin{figure}[t]
	\centering
		\includegraphics[width=0.8\columnwidth]{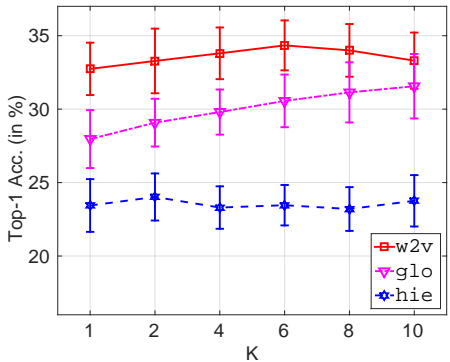}
		\caption{Effect of latent variable $K$, with unsupervised class embeddings (on CUB dataset with five splits).}
    \label{figPerfVsK}
\end{figure}

\subsection{Evaluating the number of latent embeddings}
\label{subsec:increasingk}
In Sec. \ref{subsec:soa}, when we use multiple splits of the data, although the relative performance difference
between the state of the art and our method has not changed, for some cases we observe a certain increase or 
decrease in accuracy. In this section, we investigate the experiments performed with five-folds on the CUB dataset and provide further 
analysis for a varying number of $K$. For completeness of the analysis, we also evaluate the single
matrix case, namely $K \in \{ 1,2,4,6,8,10\}$ using unsupervised embeddings, \ie \texttt{w2v}, \texttt{glo}, \texttt{hie}.

Fig.~\ref{figPerfVsK} shows the performance of the model with a different number of matrices. We observe that the performance generally increases with 
increasing $K$, initially, and then the patterns differ with different embeddings. With \texttt{w2v} the
performance keeps increasing until $K=6$ and then starts decreasing, probably due to model overfitting. With
\texttt{glo} the performance increases until $K=10$ where the final accuracy is $\approx 5\%$ higher than with $K=1$. 
With the \texttt{hie} embedding the standard errors do not increase significantly in any of the cases,
are similar for all values of $K$ and there is no clear trend in the performance.
In conclusion, the variation in performance with $K$ seems to depend of the embeddings used, however, 
in the zero-shot setting, depending on the data distribution the results may vary up to $5\%$.

\section{Conclusions}
\label{sec:conclusions}

We presented a novel latent variable based model, Latent Embeddings (\mthd), for learning a
nonlinear (piecewise linear) compatibility function for the task of zero-shot classification. \mthd
is a multi-modal method: it uses images and class-level side-information either collected through
human annotation or in an unsupervised way from a large text corpus.  \mthd incorporates multiple
linear compatibility units and allows each image to choose one of them -- such choices being the
latent variables. We proposed a ranking based objective to learn the model using an efficient and
scalable SGD based solver. 

We empirically validated our model on three challenging benchmark datasets for zero-shot
classification of Birds, Dogs and Animals.  We improved the state-of-the-art for zero-shot learning
using unsupervised class embeddings on AWA up to $66.2\%$ (vs $60.1\%$ )and on two fine-grained
datasets, achieving $34.9\%$ accuracy (vs $29.9\%$) on CUB as well as achieving $36.3\%$ accuracy (vs
$35.1\%$) on Dogs with word2vec. On AWA, we also improve the accuracy obtained with supervised
class embeddings, obtaining $76.1\%$ (vs $73.9\%$).
This demonstrates quantitatively that our method learns a latent
structure in the embedding space through multiple matrices. Moreover, we made a qualitative analysis
on our results and showed that the latent embeddings learned with our method leads to visual
consistencies.  Our stability analysis on five dataset folds for all three benchmark datasets showed
that our method can generalize well and does not overfit to the current dataset splits. We proposed
a new method for selecting the number of latent variables automatically from the data. Such pruning
based method speeds the training up and leads to models with competitive space-time complexities \cf
the cross-validation based method.

{\small
\bibliographystyle{ieee}
\bibliography{biblio}
}

\onecolumn

\section{Appendix}

Here, we provide further quantitative and qualitative results as well as more analysis on our proposed LatEm model.

\paragraph{Comparison With The State-Of-The-Art}

In this section, we provide more analysis with~\cite{LNH13} and quantitative comparisons with \cite{romera2015embarrassingly} and \cite{SGSBMN13} which are among the most relevant related work to ours. \cite{LNH13} proposes a two-step method that follows a different principle than ours: (1) Learning attribute classifiers and (2) combining the scores of these attribute classifiers to make a class prediction. Typically, the positive/negative samples to train an attribute classifier are obtained by binarizing the class-attribute matrix through thresholding. It is not clear how to extend this idea to unsupervised class embeddings, therefore, we compare~\cite{LNH13} and LatEm using attributes and our image embeddings on AWA. We obtain 56.2\% accuracy with~\cite{LNH13}, and $71.9\%$ accuracy with our LatEm model. We would like to emphasize that we focus on unsupervised embeddings which may not be easily employed in~\cite{LNH13}.

We re-implemented~\cite{romera2015embarrassingly} following the paper because their method is embrassingly simple. In~\cite{romera2015embarrassingly}, they define a binary matrix $Y$ of size $m\times z$ to denote the groundtruth labels of $m$ training instances belonging to any of the $z$ classes. The scale of this matrix has been given as $Y\in \{-1,1\}^{m\times z}$ in~\cite{romera2015embarrassingly}. As shown in Tab. \ref{tab:cub-awa-dog}, we denote the results with $Y\in \{-1,1\}^{m\times z}$ as~\cite{romera2015embarrassingly} and obtain significantly lower results than our LatEm method with the default $Y$. On the other hand, considering that the scale of matrix $Y$
is another parameter to tune, we also validated the results with $Y\in \{0,1\}^{m\times z}$. We denote the experiment that uses $Y\in \{0,1\}^{m\times z}$ in~\cite{romera2015embarrassingly} as ~\cite{romera2015embarrassingly}* in Tab. \ref{tab:cub-awa-dog}. We observe a significant increase in accuracy by changing the scaling factor as such. However, even with $Y\in \{0,1\}^{m\times z}$, our LatEm still outperforms~\cite{romera2015embarrassingly}* in 8 out of 11 cases. For~\cite{SGSBMN13}, we got the code from the authors and then we repeated the experiments as described in section 5. As it can be seen from Tab~\ref{tab:cub-awa-dog}, our LatEm consistently outperforms~\cite{SGSBMN13} on all the datasets.   

\begin{table}[h]
    \begin{center}
        \begin{tabular}{|c||c|c|c|c|c|c|c|c|c|c|c|c|}
            \hline
            & \multicolumn{4}{c|}{CUB} & \multicolumn{4}{c|}{AWA} & \multicolumn{4}{c|}{Dogs} \\ 
            & \cite{romera2015embarrassingly} & \cite{romera2015embarrassingly}* & \cite{SGSBMN13} & LatEm   & \cite{romera2015embarrassingly} & \cite{romera2015embarrassingly}* & \cite{SGSBMN13} & LatEm & \cite{romera2015embarrassingly} & \cite{romera2015embarrassingly}* & \cite{SGSBMN13} & LatEm \\ \hline \hline
        \texttt{att} &  $30.5$ & $\mathbf{47.1}$ & $29.4$ & $45.5$  & $65.3$ & $68.8$ & $54.9$ & $ \mathbf{71.9} $  & N/A &N/A    &  N/A  & N/A \\ \hline
        \texttt{w2v} &  $23.7$ & $\mathbf{33.7}$ & $24.8$ & $31.8$ &  $29.3$  & $57.4$ & $46.6$ &  $ \mathbf{61.1} $ & $10.0$ & $21.6$ & $13.7$ & $ \mathbf{22.6}$ \\ \hline
        \texttt{glo} &  $7.1$ & $\mathbf{33.3}$ & $25.8$ & $32.5$ &  $38.4$ & $61.7$ & $47.6$ & $ \mathbf{62.9} $ & $6.5$   & $20.0$ & $16.7$ &$ \mathbf{20.9}$  \\ \hline
        \texttt{hie} &  $2.1$ & $23.2$ & $17.9$ & $\mathbf{24.2}$ &  $52.2$ & $55.1$ & $40.1$ & $ \mathbf{57.5} $ & $21.3$   & $22.1$ & $14.8$ & $ \mathbf{25.2}$ \\ \hline
        \end{tabular}
        \vspace{2mm}
       \caption{Average per-class top-1 accuracy in zero-shot setting on CUB, AWA and Dogs datasets. We compare~\cite{romera2015embarrassingly}, ~\cite{romera2015embarrassingly}*, ~\cite{SGSBMN13} and Ours (Latent embedding model. $K$ is cross-validated). We use the same data partitioning as the results reported in Table 2 of our main paper. }
%\vspace{-3mm}
      \label{tab:cub-awa-dog}
      \end{center}       
\end{table}

\paragraph{Results with Combination of Embeddings}
Here, we provide results with direct comparison with~\cite{ARWLS15} where class embeddings are combined through early fusion (\emph{cnc}) and late fusion of compatibility scores calculated by averaging the scores obtained with different class embeddings (\emph{cmb}). We use the same combination of class embeddings as~\cite{ARWLS15} for fair comparison.

\newcommand{\vv}{\checkmark}
\begin{table*}[h]

 \begin{center}
 \resizebox{\columnwidth}{!} {
  \begin{tabular}{|c c c c | c c c c| c cc c | c cc c |}
	\hline
	& & & & \multicolumn{4}{c|}{AWA} & \multicolumn{4}{c|}{CUB} & \multicolumn{4}{c|}{Dogs} \\
	\hline
	\multirow{2}{*}{\texttt{att}} & \multirow{2}{*}{\texttt{w2v}} & \multirow{2}{*}{\texttt{glo}} & \multirow{2}{*}{\texttt{hie}} & \multicolumn{2}{c|}{\emph{cnc}} &  \multicolumn{2}{c|}{\emph{cmb}} &  \multicolumn{2}{c|}{\emph{cnc}} &  \multicolumn{2}{c|}{\emph{cmb}} &  \multicolumn{2}{c|}{\emph{cnc}} &  \multicolumn{2}{c|}{\emph{cmb}}  \\
	 &  &  &  & SJE & LatEm &  SJE & LatEm &  SJE & LatEm &  SJE & LatEm &  SJE & LatEm &  SJE & LatEm  \\
	 \hline

	$\vv$ & $\vv$ &       & $\vv$& $71.3$ &$64.5$ & $73.5$ &$73.6$ & $ 45.1 $&$42.0$ & $51.0 $&$46.2$ & N/A & N/A & N/A & N/A  \\
	$\vv$ &       & $\vv$ & $\vv$& $73.3$ & $70.7$& $ 73.9 $& $\mathbf{75.7}$ & $ 42.2 $& $39.7$ & $ \textbf{51.7} $& $46.6$ & N/A & N/A & N/A & N/A  \\
	\hline
	  & $\vv$ &       & $\vv$ & $53.9$ & $59.7$ & $55.5$ & $62.2$ & $28.2$ &$30.7$& $29.4$ &$\mathbf{33.2}$ & $23.5$ &$30.0$ & $26.6$ &$\mathbf{33.8}$ \\ 
	  &       & $\vv$ & $\vv$ & $60.1$ & $\mathbf{71.1}$ & $59.5$& $64.8$ & $28.5$& $31.3$ & $29.9$& $32.6$ & $23.5$& $25.9$ & $26.7$& $26.8$ \\ 
	
	\hline
  \end{tabular}
  }
\end{center}

\caption{Class embeddings combined (direct comparison with~\cite{ARWLS15}). \emph{cnc}: early fusion of class embeddings. \emph{cmb}: late fusion of scores. Bottom part shows combination results of unsupervised class embeddings and top part integrates attributes to the unsupervised class embeddings. The conclusions are the same as the main paper,\ie LatEm improves over the state-of-the-art (SJE~\cite{ARWLS15}) in all cases for unsupervised class embeddings consistently for all datasets, in almost all the cases when attributes are included in the combination.}
\end{table*}

\begin{figure*}[h!]
   \includegraphics[width=\linewidth]{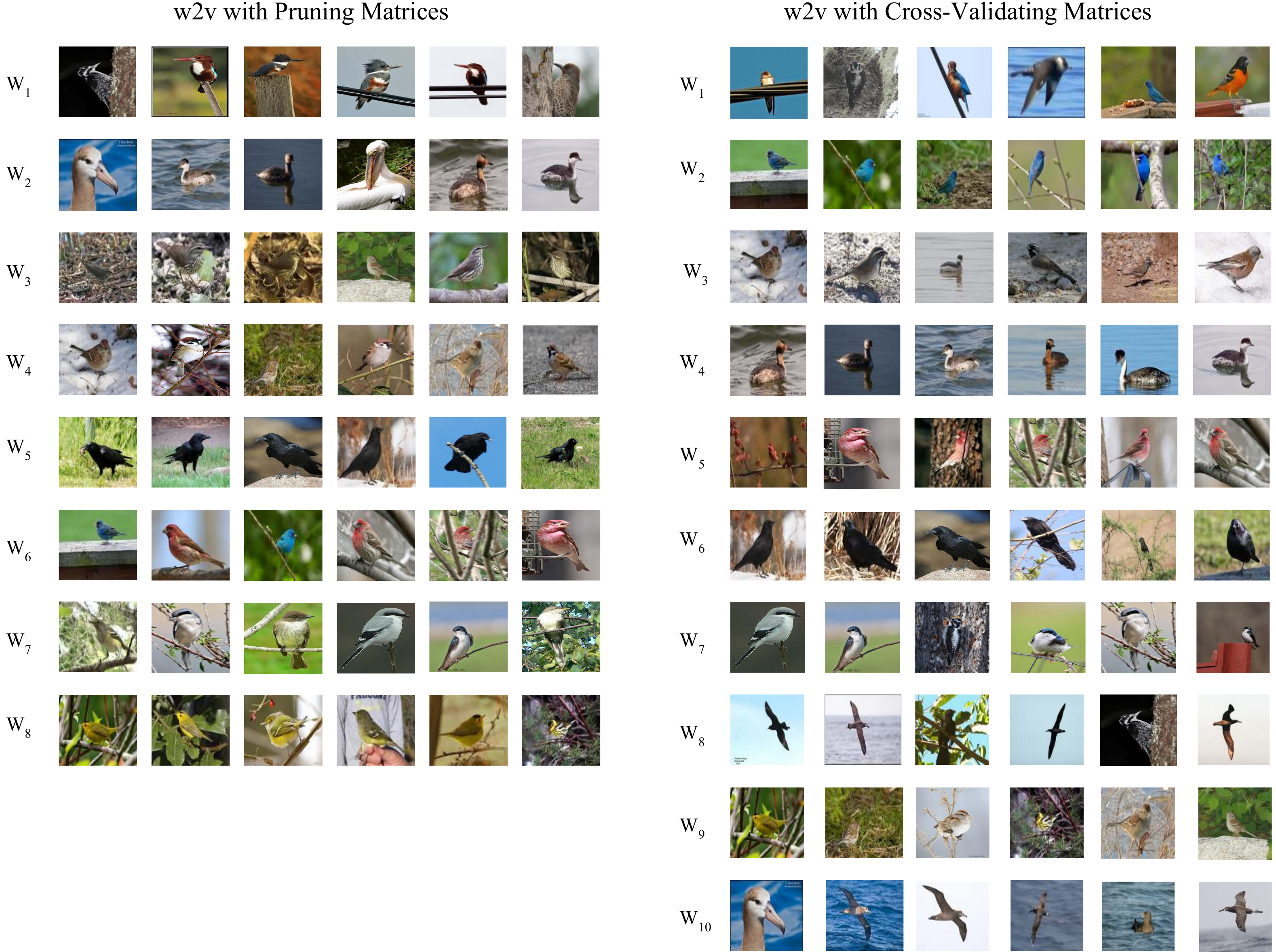}
\caption{Qualitative results with Latent Embeddings (LatEm) using \texttt{w2v} embeddings showing the highest scoring images retrieved using all learned latent embeddings $W_i$. (Left: Results with pruning, Right: Results with cross-validation.)}
\label{fig:w2v}
\end{figure*}

\begin{figure*}[h!]
   \includegraphics[width=\linewidth]{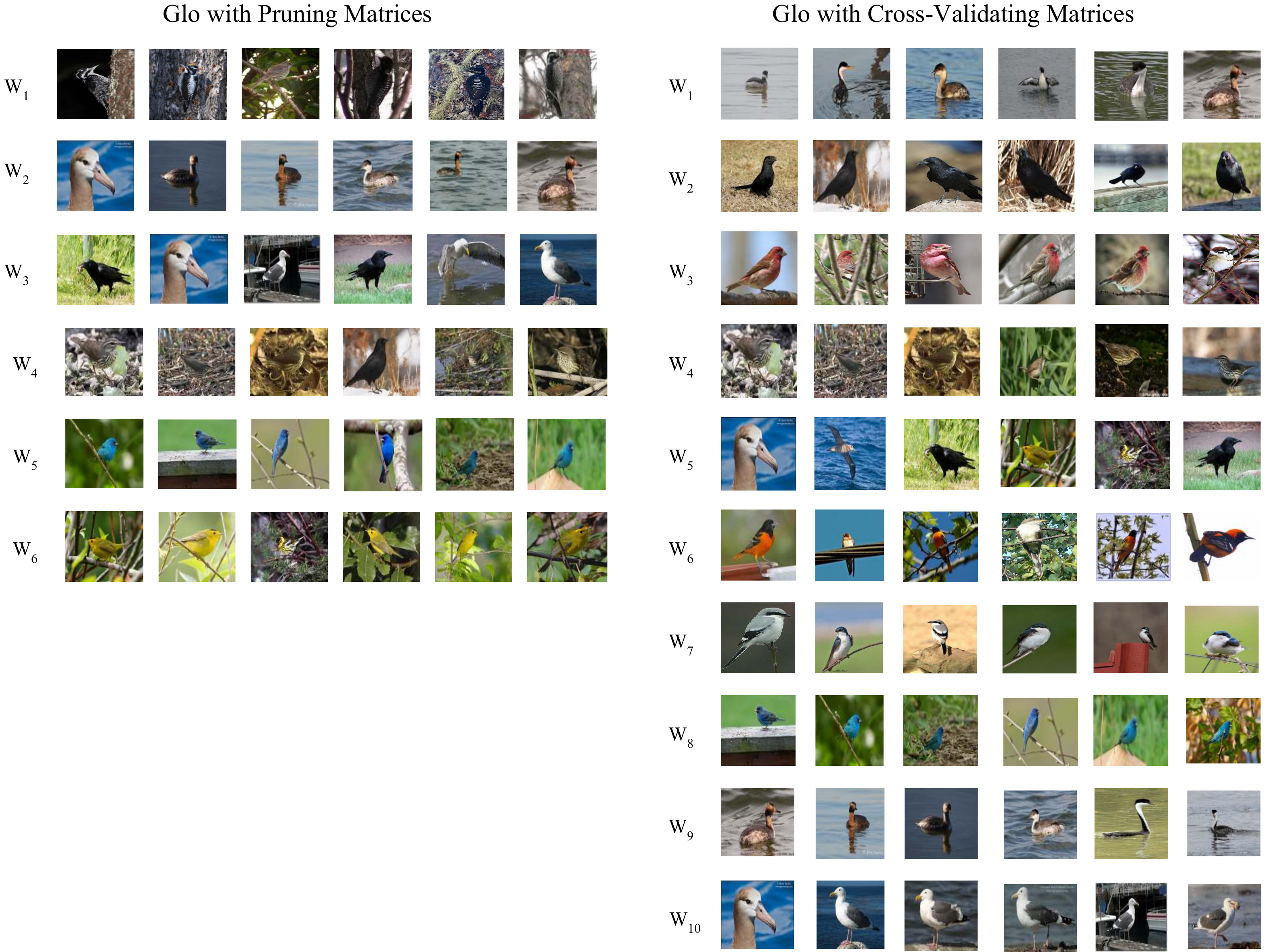}
\caption{Qualitative results with Latent Embeddings (LatEm) using \texttt{glo} embeddings showing the highest scoring images retrieved using all learned latent embeddings $W_i$. (Left: Results with pruning, Right: Results with cross-validation.)}
\label{fig:glo}
\end{figure*}

\begin{figure*}[h!]
   \includegraphics[width=\linewidth]{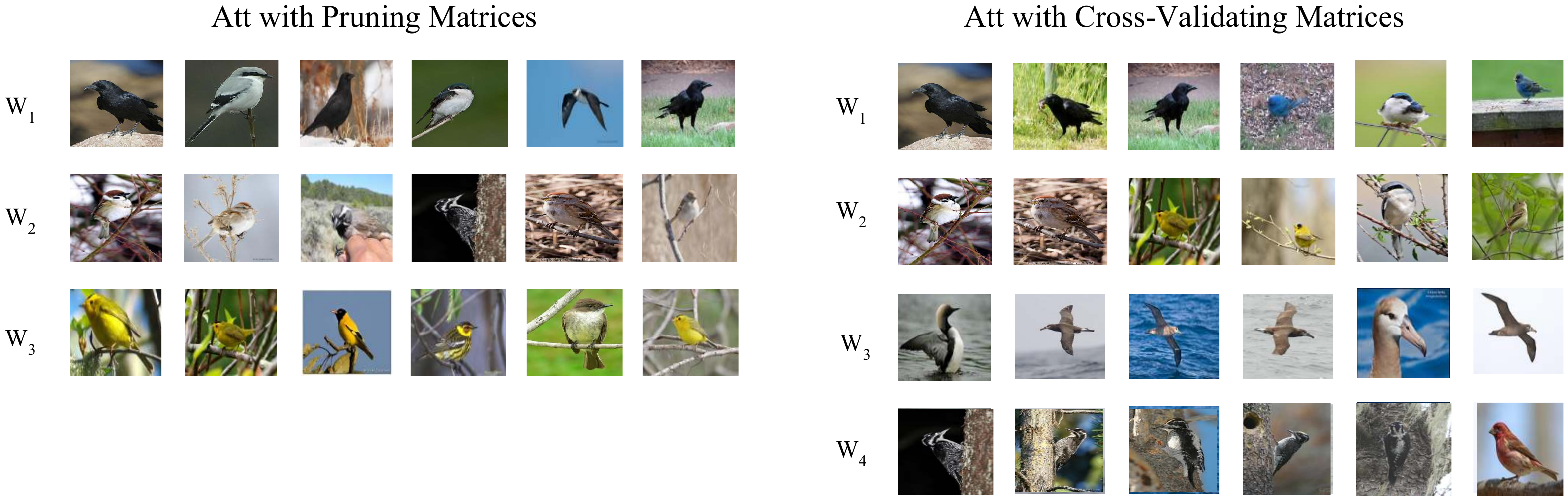}
\caption{Qualitative results with Latent Embeddings (LatEm) using \texttt{att} embeddings showing the highest scoring images retrieved using all learned latent embeddings $W_i$. (Left: Results with pruning, Right: Results with cross-validation.)}
\label{fig:att}
\end{figure*}

\begin{figure*}[h!]
   \includegraphics[width=\linewidth]{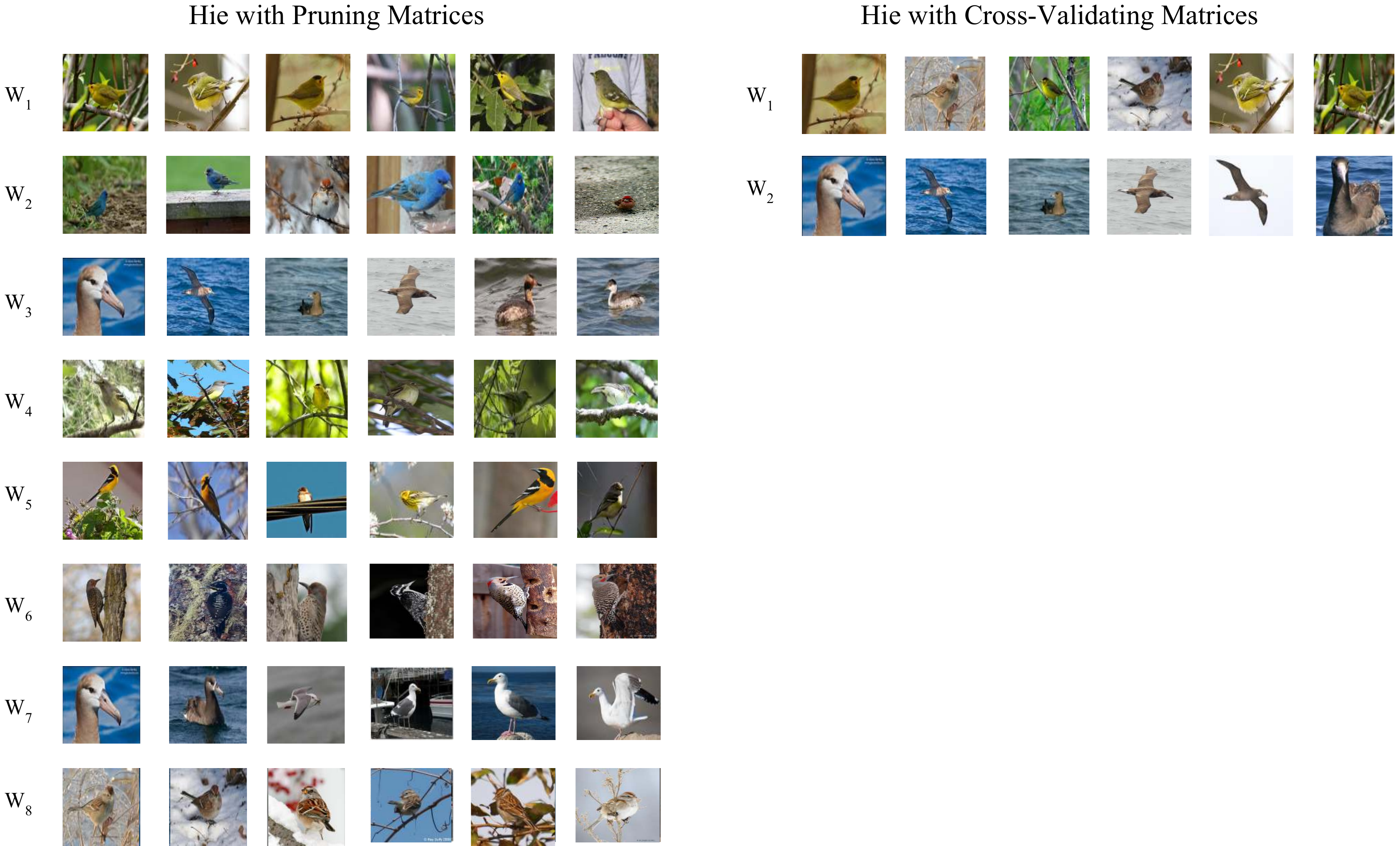}
\caption{Qualitative results with Latent Embeddings (LatEm) using \texttt{hie} embeddings showing the highest scoring images retrieved using all learned latent embeddings $W_i$. (Left: Results with pruning, Right: Results with cross-validation.)}
\label{fig:hie}
\end{figure*}

\paragraph{Further Qualitative Results}
Finally, we provide qualitative results with Latent Embeddings (LatEm) using \texttt{w2v} embeddings on Fig~\ref{fig:w2v}, using \texttt{glo} embeddings on Fig~\ref{fig:glo}, using \texttt{att} embeddings on Fig~\ref{fig:att} and using \texttt{hie} embeddings on Fig~\ref{fig:hie}. We show the highest scoring images retrieved using all learned latent embeddings $W_i$.


\end{document}